# Automatic Taxonomy Extraction from Query Logs with No Additional Sources of Information

Fernandez-Fernandez, Miguel
Tuenti Technologies
miguelfernandez@tuenti.com

Gayo-Avello, Daniel
Universidad de Oviedo
dani@uniovi.es

Abstract: Search engine logs store detailed information on Web users interactions. Thus, as more and more people use search engines on a daily basis, important trails of users common knowledge are being recorded in those files. Previous research has shown that it is possible to extract concept taxonomies from full text documents, while other scholars have proposed methods to obtain similar queries from query logs. We propose a mixture of both lines of research, that is, mining query logs not to find related queries nor query hierarchies, but actual term taxonomies that could be used to improve search engine effectiveness and efficiency. As a result, in this study we have developed a method that combines lexical heuristics with a supervised classification model to successfully extract hyponymy relations from specialization search patterns revealed from log missions, with no additional sources of information, and in a language independent way.

## 1   Introduction

Web search is becoming a common habit among internet users (Fallows, 2008). Hence, the amount of data in query logs is firmly increasing every day, recording a great deal of users' common knowledge and interactions. As (Pasca and Van Durme, 2007) pointed out *"If knowledge is generally prominent or relevant, people will (eventually) ask about it"*. Nevertheless, searching is not a straightforward process, instead, the users gradually refine both their queries and their goals in a process referred by (Spink, 1998) as the successive search phenomenon. During this iterative process the users provide successive queries revealing different search patterns (Boldi et al., 2009). The most relevant ones for this proposal are the so-called Specialization pattern; and its antisymmetric operation, the Generalization pattern. Given a pair of queries ($q_i$, $q_{i+1}$), Specialization occurs when a new query $q_{i+1}$ is focused on increasing the precision of the previous query $q_i$. In some cases a specialization can be automatically identified because $q_i$ is a substring of $q_{i+1}$ (e.g. video game sales and arcade videogame sales), in others, some of the terms are shared among both queries and the difference is more specific in $q_{i+1}$ (e.g. bird food and canary food); a specialization can even occur when none of the terms are shared among the query (e.g.: outdoor activities and camping). Generalization comes to increase the previous query recall by looking for more generic information (e.g.: white-water rafting and extreme sports).

It must be noticed that when considering groups of queries we are not interested in all the queries issued by a user during one *sitting* (i.e. a searching episode) but in much shorter fragments where all the queries are topically related. The advantages of using such topical sessions are two-fold: (1) the data to be considered in order to find semantic relations between terms is much more focused on, and (2) such granularity level should dispel most of the privacy issues even if no de-identification was used (Xiong and Agichtein, 2007). In order to obtain such query log segmentation we have used a technique which has proved to attain similar results to those achieved by a human expert (Gayo-Avello, 2009). Such technique allows us to group topically related queries even when those queries do not share any common term.

## 2 Motivation

Taxonomies are made up of terms connected by hyponymy relations. The deductive power of hyponymy allows the application of reasoning schemas based upon structural subsumption (Baader, 2003), and hence, by using taxonomies it should be possible to greatly improve search engine effectiveness and efficiency by means of term disambiguation, and semantic query suggestion and expansion. For these same purposes, other lexical databases such as Wordnet could be applied (e.g. WordNet (Miller, 1990)) but we feel they present several lacks in order to be really useful. First, because WordNet is an English language project, parallel projects for other languages have been developed, such as EuroWordNet (Ellman, 2003), BalkaNet (Greek), Hebrew WordNet, Hindi WordNet and Japanese WordNet among others (Vossen and Fellbaum, 2004). Certainly we could rely on such different wordnets but the task of identifying the language in which queries are written is not trivial given the small number of terms usually employed. Additionally, there exist a huge gap between the lexicon used by Web users and the developers of wordnets. For instance, (Mandala et al., 1999) and (Gabrilovich and Markovitch, 2007) pointed out that most domain-specific relationships between words cannot be found in WordNet, and some kind of words, such as proper names, jargon or slang are just not included. Besides, (Mihalcea, 2003) also explained that due to the fact that professional linguists recognize minimal differences in word senses, common words such as "make" have too many different senses to be useful for IR tasks. Of course, these wordnets could be automatically enriched (Hearst, 1992) but such approach require a great effort (usually carried out by linguists) and, hence, wordnets remain as quite static data sources. On the other hand, most of the previous studies aimed at building term taxonomies – e.g. (Hearst, 1992), (Berland and Charniak, 1999), (Caraballo, 1999), (Girju et al., 2003), (Morin and Jacquemin, 2004). not only need large text corpora but they are also tightly coupled to the grammar rules of the target language. This would make their application to query logs extremely difficult (if not totally unfeasible) given the nature of the queries which are short and, many times, simply ungrammatical. Thus, we feel that taxonomies of terms and noun phrases collecting the common knowledge of search engine users, including typos, jargon and slang are a real need in order to improve the performance of Web search engines. Besides, we think that the only way to obtain such users' mental model is by mining the query logs collecting the users queries. As a consequence, the following research questions come to light:

(q1) Is it possible to automatically generate term taxonomies containing the common vocabulary employed by web users?

(q2) If it is so, can it be done by using just the information contained in the query log, with no additional sources of information?

(q3) Is it possible to do all the above in a language-independent way?

Throughout the following sections we describe our method to mine hyponymy relations from query logs, how we have applied it to the AOL and MSN datasets, the results we have obtained from its application, and finally, the implications and future lines of research of the study.

## 3 Prior work

The idea depicted in this paper is somehow related to previous and on-going works. We will briefly review those which are most relevant, then, we will point out the main differences between such works and our approach. First, it must be said that the idea of automatically building term taxonomies is not new and several approaches have already been proposed to work on full text documents. Works such as (Hearst, 1992), (Berland and Charniak, 1999), (Caraballo, 1999), (Girju et al., 2003), (Morin and Jacquemin, 2004), among others, are extremely relevant but they cannot be straightforwardly applied to query logs, because most of these techniques require lexico-syntactic patterns and POS tagging which are hardly useful when applied to Web search queries.

With regards to those works relying on query logs or folksonomies, there have been two main goals: (1) organizing the queries/tags in hierarchical arrangements (but not actual taxonomies), and (2) automatically obtaining similar queries/tags.

Thus, (Clough et al., 2005) and (Schmitz, 2006) applied subsumption to image tags in order to obtain tag hierarchies. Such hierarchies, however, were not taxonomies because no hyponymy relations were established; instead, the tags were arranged with regards to their specificity (e.g. church ← tower ← bell tower, sanfrancisco ← goldengate). (Heymann and Garcia-Molina, 2006), (Mika, 2005), and (Schwarzkopf et al., 2007) developed rather similar works; they also employed tag collections (although not image tags) and described different techniques to obtain concept hierarchies. Again, such hierarchies were not proper taxonomies. With regards to the field of query suggestion there exist abundant literature; we will just refer to two recent works that could be confused with our proposal. For instance, (Shen et al., 2007) and (Baeza-Yates and Tiberi, 2007) describe two methods to generate related queries for a given one by exploiting the data within the query log; however, neither of such methods produces a proper taxonomy the way we suggest.

Approaches by other authors could be wrongly considered similar to our approach. For instance, (Chuang and Chien, 2003) describe a method to classify query terms into a predefined category system; thus, it is much closer to query topic classification than to taxonomy bootstrapping. Other works by the same authors such as (Chuang and Chien, 2004) and (Chuang and Chien, 2005), describe methods to obtain term hierarchies but such hierarchies are, in fact, clusters and not taxonomies. There also exist interesting works in the field of information extraction. For instance, (Pasca et al., 2006) and (Pasca, 2007) describe a technique to obtain class attributes from query logs (e.g. finding that population, flag or president are attributes for Country). The same author also provides a method to find named-entities (Pasca and Van Durme, 2007) which is related to (Sekine and Suzuki, 2007) and (Komachi and Suzuki, 2008). None of these works, however, are related to our approach because they do not generate term taxonomies.

Thus, our proposal, although somehow related to all the aforementioned research is different in several aspects. Different from classic works – e.g. (Hearst, 1992), (Berland and Charniak, 1999), (Caraballo, 1999), (Girju et al., 2003) and (Morin and Jacquemin, 2004) in that it does not rely on full text documents but on query logs. It also differs from (Clough et al., 2005), (Heymann and Garcia-Molina, 2006), (Schmitz, 2006), (Mika, 2005), (Baeza-Yates and Tiberi, 2007) and (Schwarzkopf et al., 2007) in the underlying goal: while those methods obtain tag or query hierarchies according to their specificity, we are interested in automatically building actual taxonomies (i.e. hierarchical arrangements according to hyponymy relations). We have also exposed that other works such as (Chuang and Chien, 2003), (Chuang and Chien, 2004), (Chuang and Chien, 2005), (Pasca et al., 2006), (Pasca and Van Durme, 2007), (Pasca, 2007), (Sekine and Suzuki, 2007) and (Komachi and Suzuki, 2008) are in fact dealing with problems which are totally unrelated to taxonomy construction.

# 4 Method applied

Our method relies on extracting pairs of terms or noun phrases from a series of query specialization patterns identified from topical query sessions. Broadly, the outline of this process consists of the following three activities:

1. Sessionize and filter the log obtaining sets of non-navigational queries targeted at solving a particular information need.

2. Identify query pairs revealing query specialization/generalization patterns from that sessions.

3. For each of the above pairs, point out a group of hyponymy candidates over which, at most, one instance will be chosen as a true hyponymy relation.

During this work, we have developed two lines of experiments. The former was presented in (Fernandez-Fernandez and Gayo-Avello, 2009) and implemented the identification of specialization search patterns by only taking into account lexico-syntactic aspects about the queries – mainly, addition and subtraction of terms. The second one, evolved from the need of increasing the method's performance, applies a supervised method devised by Bonchi *et al.* (Boldi et al., 2009) which takes into account the lexical and temporal features of query pairs, in addition to session-related information. Both lines of research are complementary: On the one side, lexical identification performs well in cases in which both queries of a pair share some terms – e.g. wild animal photographs and lion photographs, or naked celebrities and naked angelina jolie – yet it does not allow taxonomy extraction over query pairs that do not have any term in common, such as golden globe and film awards, that can be discovered using machine learning.

## 4.1 Sessionization and navigational query removal

### 4.1.1 Topical session detection

Topical sessions, or missions (Boldi et al., 2009) are sets of queries submitted by the same user pursuing the resolution of a single information need. There exist a wide range of studies describing methods to reveal topical sessions from query logs. The work by Gayo-Avello (Gayo-Avello, 2009) surveys the state of the art on this field, and describes a new method that takes into account both the lexical and temporal dimensions of a pair of queries, in order to determine whether they belong to the same session or not. This method, called **Geometric**, performs better than the others surveyed by the author ($F_{1.5} = 0.82$), and hence it has been the one chosen to sessionized our data sets.

Provided the character 3-gram vectors of a pair of queries, and their submission timestamps, our implementation calculates the (*x, y*) coordinates for the 2-dimensional space characterized by *x*) the lexical resemblance between both queries (cosine); and *y*) their temporal similarity, linearly normalized in the interval [0,1], given a maximum timespan of half-an-hour. Membership of the same session will be determined if the point (x,y) overtakes the boundaries of a circle with the center in (1,1) and radius being 1. (Figure 1).

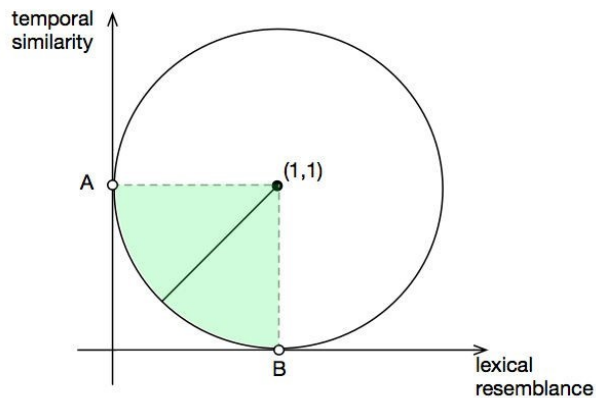

Figure 1: Geometric interpretation of the lexical and temporal dimensions of a pair of queries. Queries completely different but submitted at the same time (A), and similar queries submitted with at least half-an-hour of difference (B), belong to the same session.

### 4.1.2 Removal of navigational queries

According to (Broder, 2002), there are three kinds of queries in relation to their intent: (1) Navigational, when the immediate intent is to reach a particular site; (2) Informational, when the intent is to acquire some information assumed to be present on one or more web pages; and (3) Transactional, when the intent is to perform some web-mediated activity (e.g. to buy a product, or download a file). We think that navigational queries can reduce the accuracy of the hyponym extraction process and, thus, such queries should be removed. Because the intent behind navigational queries is to reach a particular site, most of them are lexically similar to the URL of the referred site and so, a simple heuristic to detect them (Jansen et al., 2008) consists in marking a query as navigational if it matches at least one of the following criteria:

- Contains company/business/organization/people names.
- Contains domains suffixes.
- Has "Web" as the source.
- Its length (i.e., number of terms) is less than 3; and searcher clicked on the first results page.

A straightforward application of this criteria leads to wrongly classify too many queries as navigational. Because of this, we have relaxed the heuristic and filtered as navigational only those queries containing 1) well known website names (e.g. google, wikipedia, etc), 2) domain suffixes (e.g. com, net, ... , co.uk, etc), or 3) strings frequently present in URLs such as www. or http://.

## 4.2 Reformulation pattern identification

The output of the previous activity is a collection of non-navigational topically-related sets of queries. Over them we identify pairs of queries in which one of its elements asks for more precise information.

### 4.2.1 Lexical identification

A specialization occurs when a query $q'$ looks for information about the same topic as a previous query $q$, but in a more specific way. A generalization occurs when the user wants to increase search recall by reaching more relevant documents. Both patterns are antisymmetric, meaning that there exist a specialization in $(q,q')$ if, and only if $(q', q)$ make up a generalization.

In order to detect such patterns, works like (He et al., 2002) rely on the lexical similarity between both queries in such a way that one query specializes another if it adds terms to it. A trivial scenario occurs when $q$ is a substring of $q'$ (e.g. fish food and tropical fish food). A not so trivial scenario occurs when $q'$ not only adds some terms to $q$, but also removes others, as it happens in the case of the queries celebrity scandals and charly sheen scandals. This kind of specialization – we call it specialization with reformulation – can be seen as a parallel move on the session, because the second query could look for slightly different information than the former, but if we pay attention to their number of results (11M vs. 0.4M respectively) we can see that the second one is much more specific, and thus, it is subsumed by the first one. A different case is that of the pair electronic repairs and iphone repairs, whose number of results swings in the same order of magnitude (340M vs. 550M)

### 4.2.2 Supervised identification

A third kind of specialization occurs when $q$ and $q'$ do not share any term, but $q'$ looks for more specific information, as in the case of the pair outdoor activities and camping. These patterns cannot be identified by attending only to the lexical criteria described above.

In (Bonchi *et al.*, 2009), the authors describe a machine learning approach that applying a queue of binary classifiers in cascade, is able to classify a pair of queries into the following equivalence classes:

generalization (lion, wild animals); specialization (ikea furniture, corner units), error correction (califrnia, california); parallel move, when queries look for something related, but not similar (hotel in Dublin, flights to Dublin); and session shift, when both queries are not aimed at solving the same information need and hence they do not belong to the same session.

Given that our input in this task are queries that belong to the same session, and were not previously identified as either trivial specializations or specializations with reformulation, our problem is reduced to the application of only two of the binary classifiers (those targeted at identifying specialization and generalization). The moment the query does not match any of the former categories, it is discarded. (Figure 2).

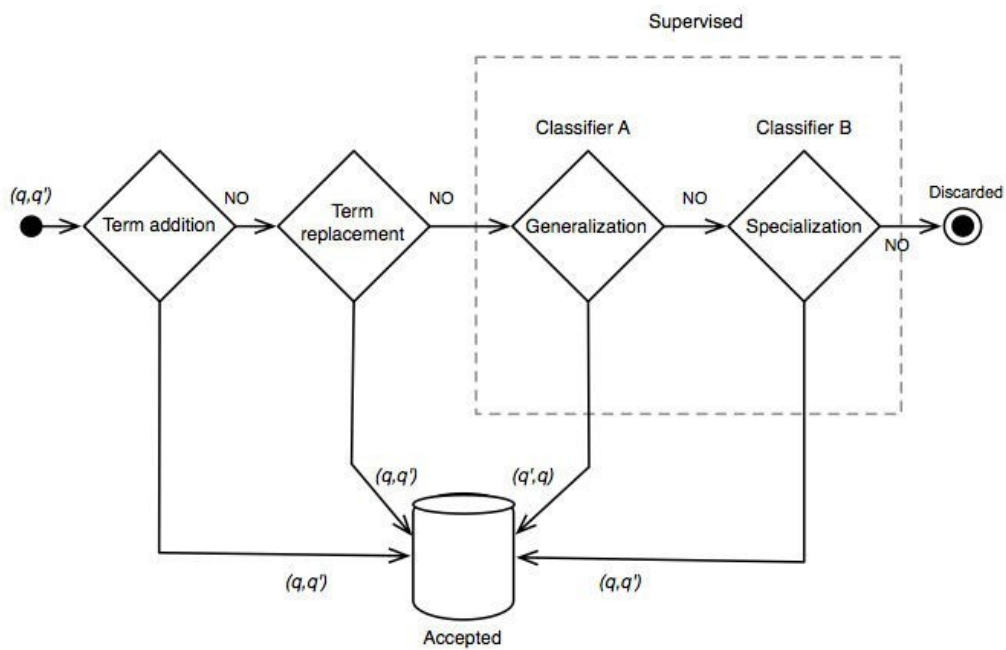

Figure 2: Flow diagram showing the decisions involved in the identification of specialization patterns.

The classification algorithm used in our work is J48, an open source implementation of the well-known C4.5 decision tree induction algorithm (Quinlan, 1993). As in the original paper, the algorithm was trained with a manually labeled sample of 3000 query pairs, but in our case, we limited them to those that 1) do not share any term, and 2) belong to the same session. The labeling process was accomplished by three judges that assigned one category among a) generalization, b) specialization and c) undefined. The result of this process was 421 query pairs labeled as either generalizations or specializations. Again, 1/3 of the labeled sample was used to evaluate the classifier's performance.

To build the model from which the trees were induced, we calculated for each query pair the 27 lexical, temporal, and session related features proposed by Bonchi *et al.* with the exception of features [f21], [f22] y [f23], that in the original work were the cosine similarity, the Jaccard coefficient, and the overlapping of the stemmed query terms. We feel that applying stemming (Porter, 1980) ties the classifier to the language in which the queries were written, so we replaced that features in favor of the same measures applied to the Soundex codes of the terms, this way we obtain a set of similarity measures that are independent of, at least, all the occidental languages.

## 4.3 Hyponymy relation extraction

The last step in the process consists in, given an specialization (*q*,*q'*), identifying which terms from *q* and *q'* act respectively as the hypernym and hyponym in the final relations. To do it, we first identify a set of candidate pairs and then we choose the most relevant candidate provided that its relevance is above a certain threshold. Relevance in this case is defined as a weight that is proportional to the probability of finding the same candidate in other specialization patterns across the log.

$$W(t,t') = \frac{|P(t,t')|^2 + 1}{2 \cdot |G(t')| \cdot |S(t)| + 1} - 1$$

In the equation, *P*(*t*, *t'*) is the subset of the specialization patterns in which the term *t* appears in the more general query, and *t'* in the more specific one; *G*(*t'*) is the subset of specialization patterns in which *t'* appears only in the more general query; and finally *S*(*t*) is the subset in which *t* appears only in the more specific query. As a consequence, on the one side, a candidate relation appearing *as it is* in more patterns, will see its weight increased in a quadratic factor. On the other side, a candidate relation whose parts appear on the remaining set of specialization patterns, playing the opposite role (the hyponym in the more general query, and the hypernym in the more specific) will see its weight reduced by a linear factor.

Depending on the kind of specialization pattern used as source, a concrete behavior is defined to select the candidate relations and, depending on the values of *W* assigned to it, to decide the instance that will be part of the taxonomy.

### 4.3.1 Harnessing specializations with reformulation

This is the easiest case. A specialization with reformulation happens when in the pair (*q*,*q'*), some of the terms of *q* are replaced by others in *q'* (e.g. naked celebrities and naked angelina jolie). In this case, the intersection between both queries is removed, and the remaining terms in *q* and *q'* are taken as hyponym and hypernym of the candidate relation. Applying this to the previous example will give the relation celebrities ← angelina jolie as a result. It is easy to figure out that not all of this kind of specializations make up such a clear relation. (e.g. president bush, president of the united states). This is the reason why the candidate must have a positive weight (*W* > 0) in order to consider it as an actual hyponymy relation.

### 4.3.2 Harnessing trivial and disjoint specializations

A trivial specialization on a pair of queries (*q*,*q'*) is that in which *q'* only adds terms to *q*, or in other words, in which *q* is a substring of *q'* (e.g. luxury cars, american luxury cars). We talk about disjoint specialization when no terms are shared between the two queries, but according to temporal, lexical and session related clues, a classifier determines that *q'* has a narrower meaning than that of *q*. (e.g. marvel superheroes, wolverine). The candidate selection heuristic is the same in both cases:

1. We compute the term n-gram vector for both queries. Following the luxury cars example, the gram vectors are *g*=[luxury, cars, luxury cars]; and *g'*=[american, luxury, cars, american luxury, luxury cars, american luxury cars].

2. The set of candidates is built by combining each of the n-grams of *g*, with every n-gram of *g'*, provided that the n-gram coming from *g* is the hypernym (*t*), the one coming from *g'* is the hyponym (*t'*), and *t'* is not a substring of *t*. (Table 1).

3. For each candidate (*t*, *t'*) we compute W.

4. Finally, we choose the candidate with the highest positive weight, but we do not choose any – and so, the extraction is considered barren – , if for every candidate $W < 0$.

| Candidate relations | W |
|---|---|
| luxury ← american | 0.66 |
| luxury ← cars | -0.24 |
| luxury ← american luxury | -0.84 |
| luxury ← luxury cars | 0.0 |
| luxury ← american luxury cars | 1.00 |
| cars ← american | 0.11 |
| cars ← luxury | 0.66 |
| cars ← american luxury | 0.66 |
| cars ← luxury cars | 4.00 |
| cars ← american luxury cars | 1.00 |
| luxury cars ← american luxury | 1.00 |
| luxury cars ← american luxury cars | 1.00 |

Table 1: Candidate pairs from the combination of the n-grams in (luxury cars and american luxury cars). The candidate cars ← luxury cars has the highest weight.

## 5 Research design

### 5.1 Datasets used

As described by (Silvestri, 2010):

> *"[…] one of the main challenges in doing research with query logs is that query logs, themselves, are very difficult to obtain."*

For this research we have used two of the latest, publicly available, logs: AOL 2006 (Pass et al., 2006) and MSN 2006 (Zhang and Moffat, 2006).

On the one side, AOL 2006 contains more than 30 million records from about 650,000 users sampled from March to May 2006. Each record in the log contains 1) a user identifier, 2) the query string submitted by the user, 3) the timestamp of the submission; and if the user clicked on any result, then the record also includes 4) the position of the result clicked, and 5) the hostname portion of the visited URL.

On the other side, the MSN 2006 log was released as part of the "Microsoft Live Labs: Accelerating Search in Academic Research" [1] incentive in 2006. This dataset contains about 15 million queries submitted by users from the United States during May 2006, as recorded by the MSN search engine. For each query, in addition to the same information provided by the AOL query log, this dataset also contains the number of search results that satisfied the query. One major difference between both logs is that, while AOL contains immutable user identifiers for every record originated by a certain user, the MSN log is anonymized in a way that user identifiers change each 30 minutes, preserving the users' privacy.

As it was previously explained, we rely on the number of results to check query subsumption and identify specialization with reformulation patterns. This data can appear in every query log (as it appears in MSN), but has been omitted from the AOL one. To recreate this information, we have resubmitted each query to the Yahoo! BOSS[2] API. And to reduce impedances between the results gathered from Yahoo!, and those present in the MSN log, we did the same operation with the latter.

At the end of this preprocessing activity, our dataset was comprised by near 45 million records containing 1) a user identifier, 2) the query string, 3) the submission timestamp, 4) the number of results of each query, and 5) click-through information.

## 5.2 Method implementation

The experiments were applied by following a pipeline architecture in which the initial dataset is transformed into new data structures (figure 3). Some of the activities in the workflow are conceived to increase the method performance in two ways: 1) increasing effectiveness by reducing informational noise in the log; and 2) improving efficiency by generating new ad-hoc data structures that favor information extraction in the specialization detection, and in the relation extraction activities.

### 5.2.1 Noise filtering

Besides navigational queries, there exist others that – because of their nature – are not valid to mine semantic relations from them. For instance, the most frequent query in the AOL log is the " " query, which is believed to be the result of a masking strategy by the search engine (Brenes and Gayo-Avello, 2009). In the MSN log, those queries belonging to the longest hundred sessions are recorded in an average of less than three seconds each, and there is no click information at all associated to them. This suggests that the queries were sent by a software agent through the search API (Zhang and Moffat, 2006).

To reduce the amount of useless information in the log, we have defined the following criteria to consider a query as spam:

- The number of characters for all the terms in the query is lower than 3.
- The number of characters in any term in the query is greater than 25.
- The number of terms in the query is greater than 5.
- The average time between query submissions from the same user is lower than 7 seconds.

---
[1] http://research.microsoft.com/ur/us/fundingopps/RFPs/Search_2006_RFP.aspx
[2] http://developer.yahoo.com/search/boss

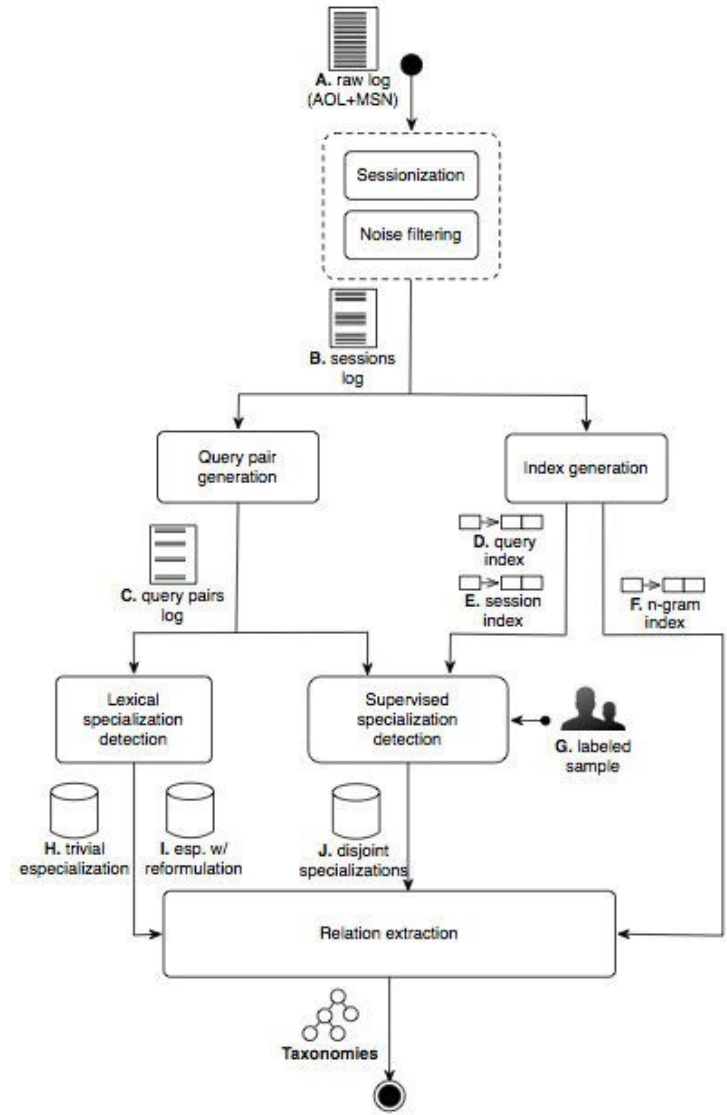

Figure 3: The experiments workflow and the products generated in each activity

A user who sends a query that matches at least one of the points above is considered a spammer, and the whole session where the query appears is invalidated. The product of this activity is in turn a set of sessions that are free of spam.

### 5.2.2 Additional data structures

Applying some of the algorithms depicted in the method require additional data structures, like indices and relations. For instance, for the supervised detection of specialization patterns, there is the need to compute session related features such as *The average number of clicks in search results since session begin, among all sessions containing (q,q′)*. For this sake, from the sessionized logs we create the following indexes:

- $q \rightarrow [r, \{s, [\{i, t\}]\}]$ given a query *q*, its number of results, the list of sessions in which *q* appears, and inside the sessions, the position in which *q* appears and the time at which is has been submitted.

- $s \rightarrow [\{q, \{p, url\}\}]$ given a session, the list of queries that make it up, and the information associated to the visited results (position and target url).

In addition, during the relation extraction activity, we need to calculate the weight assigned to each candidate. To do so, we need to know in which queries a given n-gram appears, and its length and offset inside a query ($t \rightarrow [q, p, o]$). With this information we can reenter the query index and reveal which role it plays inside a certain n-gram.

## 5.3 Proposed evaluation method

In order to measure the performance of the method developed, we must determine whether an extracted relation $t \leftarrow t'$ has hyponymy semantics or, in other words, if any of the following statements are true: *"t' is a t"*, or *"t' is an instance of t"*.

In absence of another criterion of reference, we proposed a combined technique in which first, we test if a certain relation appears as an hyponymy relation in Wordnet (and therefore it is a valid one) or, if not, we delegate to a human judgement. To support this, we rely on the following assertions:

- Most of AOL and MSN queries are written in English, the same language that Wordnet is built in.

- Based on the previous one, we can apply stemming to cushion lexical deviations of the same term.

- Some of the relations extracted capture common vocabulary of search engine users, in which is not unfrequent the use of jargon, slang, trade marks, and even typos. Although this kind of vocabulary does not appear in Wordnet, relations containing it can be identified by the judges.

Once taken all the above into account, the following evaluation process is implemented:

1. We take a sample composed by an equal number of relations extracted (E), and discarded (D) in the relation extraction activity. As the extraction algorithm varies, and we want segmented performance numbers, the sample will also have an equal number of relations extracted from each kind of specialization pattern.

2. We create a directed graph with the hyponymy relations present in Wordnet. In that graph, the vertices are the result of applying stemming over the terms that appear in the hyponymy relations, and there is an arc that goes from an hyponym to its direct hypernym.

3. Given a candidate pair, we apply stemming to each side of the relation. Then we look whether there exists a path in that graph, going from the specific side of the relation to the general side. If it does exist, then we will have finished, and the relation is correct.

4. If the candidate pair is not present in Wordnet, two judges will evaluate if it is an actual hyponymy relation, and it it is not, they will classify the candidate as either a) synonyms (co-hyponyms), b) terms related by another kind of relationship, or c) completely unrelated terms. In case that the judges don not agree on the verdict, the following rules apply:

- If both judges determine that the relation is not an hyponymy, but differ on the category assigned, the result will be c) – both terms are unrelated.
- If only one of the judges determines that the relation is a proper hyponymy, the result will be the error category assigned by the other judge.

5. The last step is to calculate precision (P) and recall (R) measures in the context of classification. P is the portion of relations properly classified by the extraction algorithm (true positives) from the whole set of relations extracted (true positives + false positives). Recall on the other side, is the portion of relations properly classified (true positives) from the whole set of existing relations (true positives + false negatives).

$$P = \frac{true\ positives}{true\ positives + false\ positives}$$

$$R = \frac{true\ positives}{true\ positives + false\ negatives}$$

# 6 Results

For each specialization type from which the relations were extracted (trivial specialization, specialization with reformulation, and disjoint specialization), we have taken 500 instances identified as valid hyponymy relations (E), and 500 instances discarded by the extraction heuristic, counting for a total of 3000 instances. Then we have applied the evaluation method described above, and obtained the following results.

|  | Wrong (false positives) | | | | Correct (true positives) | | |
| --- | --- | --- | --- | --- | --- | --- | --- |
|  | Total | co-hyponyms | other | unrelated | Total | Wordnet | Judges |
| trivial | 143 | 6 | 2 | 135 | 357 | 258 | 99 |
| w/reformulation | 119 | 11 | 3 | 105 | 381 | 259 | 122 |
| disjoint | 48 | 3 | 2 | 43 | 452 | 322 | 130 |
| Aggregated | 310 | 20 | 7 | 283 | 1190 | 839 | 351 |

Table 2: Results for the extracted subset (E). For the false positives, we display how many of them were judged as co-hyponymy relations, another – unspecified – type of relations, or as totally unrelated terms.

|  | Wrong (false negatives) | | | Correct (true negatives) | | | |
| --- | --- | --- | --- | --- | --- | --- | --- |
|  | Total | Wordnet | Judges | Total | co-hiponyms | other | unrelated |
| trivial | 81 | 25 | 56 | 419 | 16 | 0 | 403 |
| w/reformulation | 14 | 3 | 11 | 486 | 31 | 5 | 450 |
| disjoint | 85 | 24 | 61 | 415 | 18 | 1 | 396 |
| Aggregated | 180 | 52 | 128 | 1320 | 65 | 6 | 1249 |

Table 3: Results for the sample of discarded relations (D). For those wrongly classified (actual hyponymy relations) we show how many of them were present in Wordnet in contrast to those ones determined by the judges.

|                | $|E|$ | $|D|$ | TP   | FP  | FN  | P     | R     | $F_1$ | $F_{0.5}$ |
|----------------|-------|-------|------|-----|-----|-------|-------|-------|-----------|
| trivial        | 500   | 500   | 357  | 143 | 81  | 0.714 | 0.815 | 0.761 | 0.732     |
| w/reformulation| 500   | 500   | 381  | 119 | 14  | 0.762 | 0,964 | 0.851 | 0.795     |
| disjoint       | 500   | 500   | 452  | 48  | 85  | 0,904 | 0.842 | 0.872 | 0,891     |
| Aggregated     | 1500  | 1500  | 1190 | 310 | 180 | 0,793 | 0,868 | 0.829 | 0,807     |

Table 4: Classification performance (*microaveraged*). The following data is displayed: extracted and discarded set sizes ($|E|$ and $|D|$ respectively), true positives (TP), false positives (FP), false negatives (FN), precision (P), recall (R), balanced F-score ($F_1$), and precision-emphasized F-score ($F_{0.5}$).

|                | $N_i$         | $\frac{P \cdot N_i}{N}$ | $\frac{R \cdot N_i}{N}$ | $\frac{F_1 \cdot N_i}{N}$ | $\frac{F_{0.5} \cdot N_i}{N}$ |
|----------------|---------------|-------------------------|-------------------------|---------------------------|-------------------------------|
| trivial        | 245.997       | 0.478                   | 0.546                   | 0.509                     | 0.490                         |
| w/reformulation| 117.321       | 0.243                   | 0.308                   | 0.272                     | 0.254                         |
| disjoint       | 4.177         | 0.010                   | 0.010                   | 0.010                     | 0.010                         |
| Aggregated     | $N = 367.495$ | $P = 0,731$             | $R = 0.863$             | $F_1 = 0,791$             | $F_{0.5} = 0,754$             |

Table 5: Classification performance (*macroaveraged*). For each kind of specialization pattern, proportional contribution to the global precision, recall and F-measures are displayed, in addition to total the number of specializations found on the provided datasets.

# 7 Implications and future directions

## 7.1 Discussion

From the initial dataset comprised of about 45 million queries, 367,495 were identified as specialization patterns. From these patterns, 51,300 hyponymy relations were extracted, from which only 7,714 were not repeated. This results show that only one out of 3,000 query pairs generate a valid hyponymy relation, and as a consequence a much bigger source of information is needed to effectively use this method in a large-scale search system. Despite all, a commercial search engine can serve up to 4,700 million queries a day (Comscore, 2012), and this is more than 100x the data we had for this study. We believe this vast amount of information is more than the necessary to build a useful taxonomy.

Regarding performance, it can be seen that the method behaves better when dealing with disjoint specializations ($P = 0.904$). At the same time this is the less frequent kind of specializations, with less than 1% of contribution to the whole method's performance ($P = 0.731$). This happens due to the fact that disjoint specializations do not share any term between the two queries, and as the sessionization algorithm takes into account query similarity, it is very uncommon that two dissimilar queries appear on the same topical-session, reducing the number of instances of this kind.

Another interesting aspect is that the number of false positives derived from a wrong interpretation of co-hyponymy relations (e.g. `bill clinton` ← `monica lewinsky`) is consistently higher in specializations with reformulation. This happens because the number of results for one of queries in the pattern is much higher than the other (i.e. the subsumption algorithm will determine an specialization) (figure 4), and the terms also appear in a wide variety of other patterns playing the same role (e.g. `clinton scandal`, `lewinsky scandal`) increasing the weight $w$ of the candidate.

Those relations whose terms appear as other kind of relations different from the hyponymy and co-hyponymy can be considered marginal (less than 0.5%). Other relations whose terms are determined as unrelated (91.29%) appear in a proportional way for each specialization type, and are caused by the existence of stop words such as from, to or is, and also by others with an extremely high frequency (e.g. britney spears), which appear in different contexts increasing candidates' weight. In order to cushion the impact of such terms, we could integrate *tf x idf* measures into the weight calculus, thus reducing the overall score of a relation containing less relevant terms. This is left for future work.

Finally, we can highlight the fact that a consistent ratio of the relations (29.6%) does not exist in Wordnet. Wordnet contains 155.287 words arranged in 117.000 sunsets Miller et al. (1990), covering almost an 85% of the lexicon in the Oxford Dictionary[3]. This ratio means that nearly 3 out of 10 relations contain terms that are common in users' vocabulary, but do not appear in formal lexicons, such as brand names (e.g. briefs ← speedo); people names (celebrities ← angelina jolie), or even typos (britney spears ← brittney spears). This semantic information clearly serves to the purpose of increasing query suggestion and expansion effectiveness.

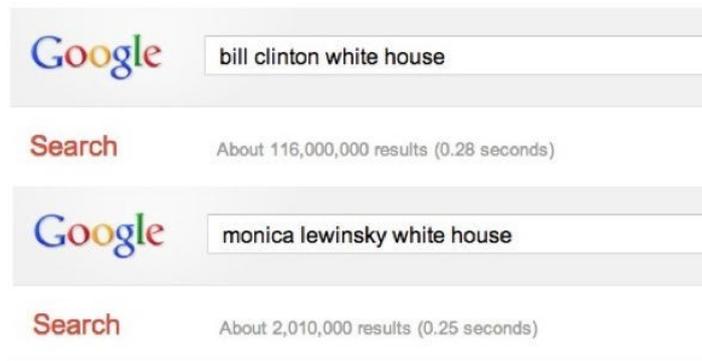

Figure 4: Number of results for clinton white house and lewinsky white house

## 7.2 Conclusions

Given the results shown and discussed above, we are now able to answer the research questions that motivated this study:

(q1) Is it possible to automatically generate term taxonomies containing the common vocabulary employed by web users?

Yes, it is. And, in addition, our method has an acceptable performance ($F_{0.5} = 0.754$), and can be implemented for large scale deployments with minimum effort.

(q2) If it is so, can it be done by using just the information contained in the query log, with no additional sources of information?

Yes, it can. The only information that is not contained in the query log is the training set used in the supervised identification activity. This information, like the rest of the code for the algorithms used, is only required at design time and does not need any additional maintenance. Other sources of information like dictionaries, search result snippets, text corpora, or any kind of semantic repositories are not used at all, so we can conclude that our method only operates on information contained in the query log.

---

[3] http://www.oup.com/online/oed/

(q3) Is it possible to do all the above in a language-independent way?

Most of the queries in our dataset are written in English. Due to this, we could not measure the performance of our method in other languages. However, we have no evidence that it cannot work for other languages in which queries can be split into terms and these, in turn, into character n-grams as this is the only lexical information actually used. There is no evidence either, that our method could not be applied to Chinese and other languages in which text is not segmented, as other scholars (Yang et al., 2000; Gao et al., 2005) have developed methods to deal with text segmentation in these languages.

To sum up, we have developed a method for the automatic extraction of hyponymy relations using query logs as the only source of information with independence of the language in which the queries are written – at least for the occidental ones – , and capturing the common vocabulary of web search users.

This will allow:

- To increase web search engines effectiveness and efficiency by improving query suggestion and expansion methods.

- To organize the *parole* of web search users in a reasonable and fast way, reflecting everyday aspects of the language that are not covered by formal classification systems, such as linguistic dictionaries and wordnets.

- To obtain term taxonomies for languages in which wordnets are scarce or do not even exist.

- To serve as inspiration for future and ongoing works on semantic information extraction, and from other limited sources of information such as folksonomies or micro-posts.

## 7.3  Future work

In the same research line, it would be interesting to:

- Try to increase the ratio of disjoint specializations by applying supervised pattern identification also for revealing topical sessions.

- Identify other kind of relationships different from hyponymy.

- Study the literature on text segmentation and entity recognition and try to apply them both to extract taxonomies in other languages, and to increase the accuracy of our method.

- Develop a model for related query recommendation.

Other related lines of research, would be:

- Enriching other existing datasources with semantic information (e.g Wikipedia, DBPedia, Freebase, etc.)

- Filtering and curating taxonomies with the semantic information contained in the previous repositories.

- Using taxonomies for semantic query tagging (i.e. determine which terms in the query describe a person, a place, or a product, among others).

- Given proper tagged queries, determine the user's intention behind each one.

- The application of the points above (semantic query tagging and identification of user intention) to filter search results, and to provide new ways of arranging them, further than document lists.

- Apply what we have learnt to new user generated data sources, such as Twitter.